\documentclass[final]{cvpr}

\usepackage{times}
\usepackage{epsfig}
\usepackage{graphicx}
\usepackage{amsmath}
\usepackage{amssymb}

\usepackage{algorithm}
\usepackage{algorithmic}
\usepackage{booktabs}
\usepackage{multirow}
\usepackage{bm}

\usepackage{float}

\usepackage[pagebackref=true,breaklinks=true,colorlinks,bookmarks=false]{hyperref}

\def\cvprPaperID{1958} 
\def\confYear{CVPR 2021}

\begin{document}

\title{Primitive Representation Learning for Scene Text Recognition}

\author{Ruijie Yan\ \ \ \ \ \ \ \ Liangrui Peng\ \ \ \ \ \ \ \ Shanyu Xiao\ \ \ \ \ \ \ \ Gang Yao\\
Beijing National Research Center for Information Science and Technology \\
Department of Electronic Engineering, Tsinghua University, Beijing, China\\
{\tt\small \{yrj17, xiaosy19, yg19\}@mails.tsinghua.edu.cn, penglr@tsinghua.edu.cn}
}

\maketitle

\pagestyle{empty} 
\thispagestyle{empty}

\begin{abstract}
   Scene text recognition is a challenging task due to diverse variations of text instances in natural scene images. Conventional methods based on CNN-RNN-CTC or encoder-decoder with attention mechanism may not fully investigate stable and efficient feature representations for multi-oriented scene texts. In this paper, we propose a primitive representation learning method that aims to exploit intrinsic representations of scene text images. We model elements in feature maps as the nodes of an undirected graph. A pooling aggregator and a weighted aggregator are proposed to learn primitive representations, which are transformed into high-level visual text representations by graph convolutional networks. A Primitive REpresentation learning Network (PREN) is constructed to use the visual text representations for parallel decoding. Furthermore, by integrating visual text representations into an encoder-decoder model with the 2D attention mechanism, we propose a framework called PREN2D to alleviate the misalignment problem in attention-based methods. Experimental results on both English and Chinese scene text recognition tasks demonstrate that PREN keeps a balance between accuracy and efficiency, while PREN2D achieves state-of-the-art performance.

\end{abstract}

\section{Introduction}

\begin{figure}[t]
\begin{center}
    \includegraphics[width=1.\linewidth]{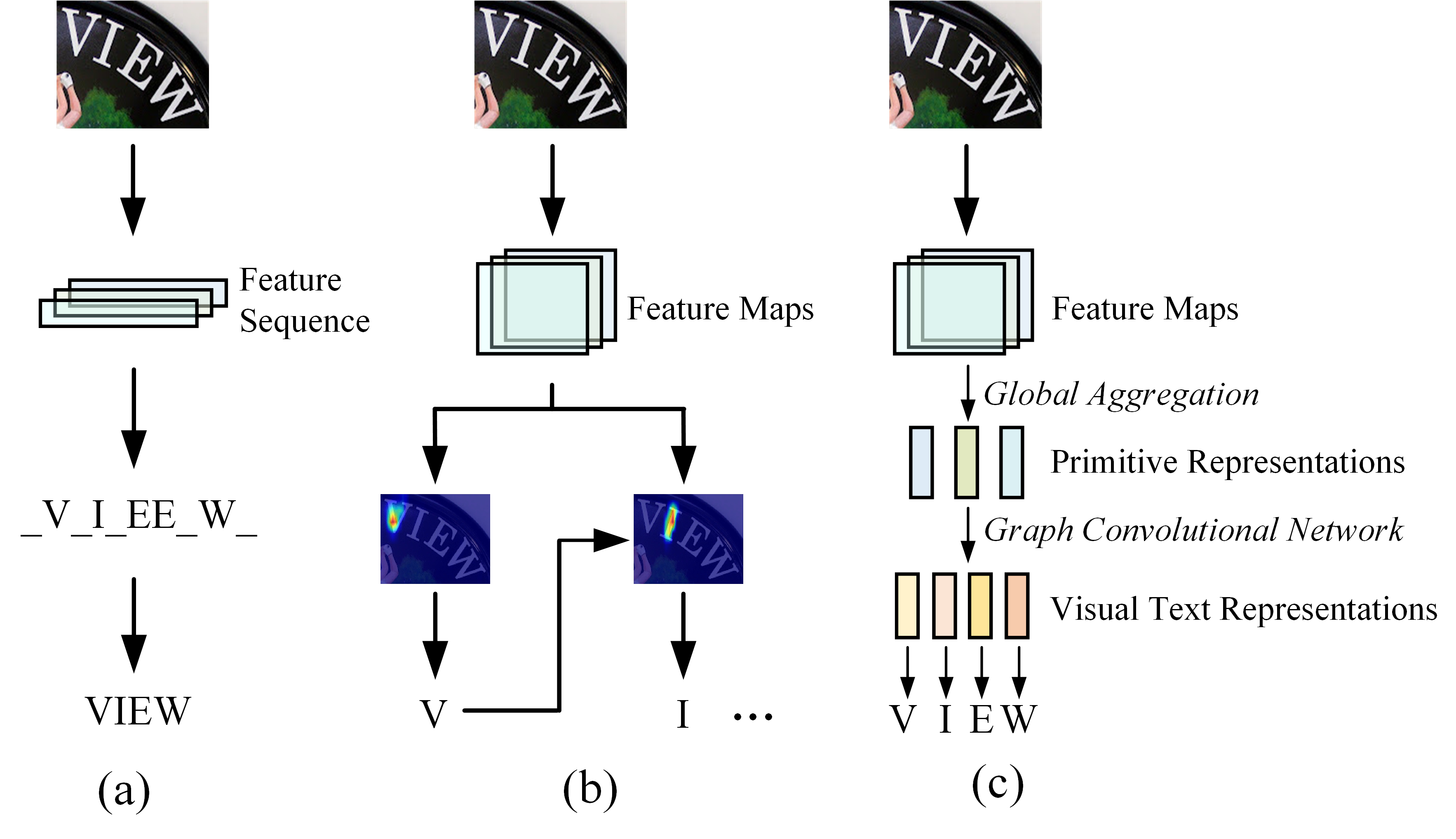}
\end{center}
   \caption{Illustrations of different scene text recognition frameworks. (a) CTC-based methods, where ``\_'' denotes the blank symbol;  (b) attention-based methods; (c) the proposed PREN.}
\label{fig1}
\end{figure}

In recent years, there have been increasing demands for scene text recognition in various real-world applications, such as image search, instant translation, and robot navigation. With the emergence of deep learning, there are two main scene text recognition frameworks. One is the CRNN framework~\cite{su2014accurate,he2015reading,pan2016reading,shi2016end,liu2016star,hu2020gtc} that encodes images into hidden representations by CNNs and RNNs, and uses the connectionist temporal classification (CTC)~\cite{graves2006connectionist} for decoding, as shown in Fig.~\ref{fig1} (a). The other is the attention-based encoder-decoder framework~\cite{bahdanau2014neural,shi2016end,lee2016recursive,cheng2017focusing,bai2018edit,liu2018squeezedtext,sheng2018nrtr,wang2019decoupled,yue2020robustscanner,litman2020scatter,qiao2020seed} that can learn to align output texts with feature maps, as shown in Fig.~\ref{fig1} (b).

However, the above methods still have room for improvement. On the one hand, for CTC-based methods, the extracted feature sequences contain redundant information that may degrade the performance on irregular text images. On the other hand, attention-based encoder-decoder methods usually suffer from the misalignment problem~\cite{cheng2017focusing,wang2019decoupled}, because the alignment between feature maps and texts is highly sensitive to previous decoded results, which lack global visual information. Therefore, to handle the diversity of texts in natural scenes, it is important to exploit intrinsic representations of scene text images.

In this paper, we propose a novel scene text recognition framework that learns primitive representations of scene text images. Inspired by graph representation learning methods~\cite{kipf2016semi,hamilton2017inductive,petar2018graph}, we model the elements in feature maps as nodes of an undirected graph. Primitive representations are learned by globally aggregating features over the coordinate space and are then projected into the visual text representation space, as shown in Fig.~\ref{fig1} (c).

The ``primitive'' representations refer to a set of base vectors that can be transformed into character-by-character vector representations in the so-called visual text representation space. The visual text representations are generated from original feature maps, which are different from character embeddings generated from ground truth or predicted texts used in an encoder-decoder model.

For the global feature aggregation, a pooling aggregator and a weighted aggregator are proposed. For the pooling aggregator, each primitive representation is learned from input feature maps through two convolutions followed by a global average pooling layer. In this way, aggregating weights are shared by all samples to learn intrinsic structural information from various scene text instances. For the weighted aggregator, input feature maps are transformed into sample-specific heatmaps, which are used as aggregating weights.

Visual text representations are generated from primitive representations by graph convolutional networks (GCNs)~\cite{kipf2016semi,chen2019graph}. Each visual text representation is used to represent a character to be recognized.

A primitive representation learning network (PREN) is constructed. PREN consists of a feature extraction module that extracts multiscale feature maps from input images and a primitive representation learning module that learns primitive representations and generates visual text representations. Texts are generated from visual text representations with parallel decoding.

Moreover, since visual text representations are purely learned from visual features, they can mitigate the misalignment problem~\cite{cheng2017focusing,wang2019decoupled} of attention-based methods. We further construct a framework called PREN2D by integrating PREN into a 2D-attention-based encoder-decoder model with a modified self-attention network.

We conduct experiments on seven public English scene text recognition datasets (IIIT5k, SVT, IC03, IC13, IC15, SVTP, and CUTE) and a subset of the RCTW Chinese scene text dataset. Experimental results show that PREN keeps a balance between accuracy and speed, while PREN2D achieves state-of-the-art model performance.

In summary, the main contributions of the paper are as follows.

\begin{itemize}
    \item Different from commonly used CTC-based and attention-based methods, we provide a novel scene text recognition framework by learning primitive representations and forming visual text representations that can be used for parallel decoding.
    \item We propose a pooling aggregator and a weighted aggregator to learn primitive representations from feature maps output by a CNN, and use GCNs to transform primitive representations into visual text representations.
    \item The proposed primitive representation learning method can be integrated into attention-based frameworks. Experimental results on both English and Chinese scene text recognition tasks demonstrate the effectiveness and efficiency of our method.
\end{itemize}

\section{Related Work}
\label{related}

\subsection{Scene text recognition}
Scene text recognition methods can be generally divided into segmentation-based methods and sequence-based methods. For segmentation-based methods~\cite{wang2011end,wang2012end,neumann2012real,yao2014strokelets,yao2014unified,lee2014region,jaderberg2014deep,zhang2016multi,liao2019scene}, individual characters are segmented or localized before recognition, and character-level annotations are often required to train these models. For sequence-based methods, CTC-based methods~\cite{graves2006connectionist,su2014accurate,he2015reading,shi2016end,liu2016star,hu2020gtc} and encoder-decoder frameworks with attention mechanisms~\cite{bahdanau2014neural,shi2016end,lee2016recursive,cheng2017focusing,bai2018edit,liu2018squeezedtext,sheng2018nrtr,wang2019decoupled,yue2020robustscanner,litman2020scatter,qiao2020seed} are two major techniques to recognize scene text images.

In contrast to CTC-based methods, attention-based encoder-decoder methods can learn the dependencies among the output characters, which can be regarded as using an implicit language model. However, the efficiency of attention-based methods is usually limited by the recursive decoding process. To increase the decoding speed while maintaining high recognition performance, Hu et al.~\cite{hu2020gtc} proposed training a CTC-based model with the guidance of an attention branch. Lyu et al.~\cite{lyu20192d} developed a two-stage decoder with a relation attention module. Yu et al.~\cite{yu2020towards} proposed a parallel visual attention module followed by a self-attention network with multi-way parallel transmission to learn semantic information explicitly. Different from these methods, we propose a novel scene text recognition framework with parallel decoding based on primitive representation learning.

Recently, the recognition of irregular scene texts has attracted a lot of research interests. The solutions include text rectification~\cite{shi2016robust,shi2018aster,luo2019moran,liu2018char,yang2019symmetry,zhan2019esir}, hierarchical attention mechanism~\cite{liu2018char}, and multidirectional feature extraction~\cite{cheng2018aon}. Models with the 2D attention mechanism~\cite{xu2015show,li2019show,lyu20192d} have also shown strong effectiveness on irregular text recognition by retaining 2D spatial information in features. Our proposed primitive representation learning method can be integrated into 2D-attention-based frameworks to improve recognition performance.

\begin{figure*}[t]
\centering
\includegraphics[width=.9\linewidth]{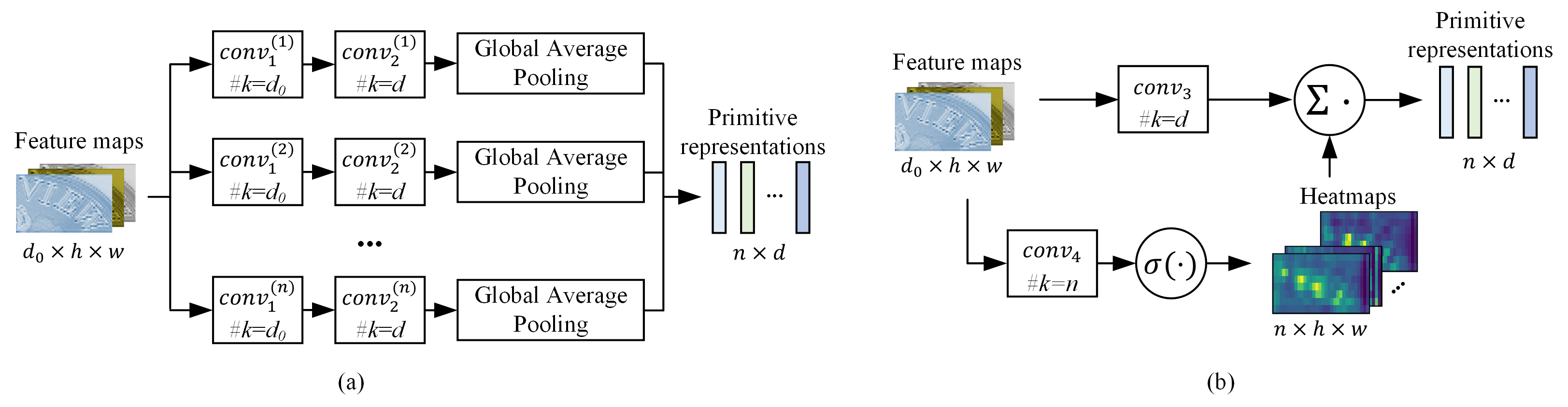}
\caption{Two primitive representation learning methods. (a) Pooling aggregator. Two convolutional layers followed by a global average pooling layer aggregates input feature maps into a primitive representation. \#$k$ denotes the number of kernels of the convolutional layer. (b) Weighted aggregator. Input feature maps are transformed into $n$ heatmaps. Each heatmap is used as aggregating weights of a primitive representation. $\sigma(\cdot)$ is the sigmoid activation function, and $\Sigma \cdot$ refers to scaled-dot product and summation.}
\label{fig2}
\end{figure*}

\subsection{Representation learning by feature aggregation}

Representation learning has become the basis of most deep-learning methods due to its ability to learn data representations that make it easier to extract useful information when building classifiers or other predictors~\cite{bengio2013representation}. Feature aggregation is a commonly used method in graph representation learning tasks. GCNs~\cite{kipf2016semi,chen2019graph} aggregate neighboring vertex features by exploiting the graph topology. Instead of using all neighboring nodes, GraphSAGE~\cite{hamilton2017inductive} uses random walk~\cite{bryan2014deepwalk} to sample several neighboring nodes, and the feature aggregation can be accomplished by a mean aggregator, a pooling aggregator, or an LSTM aggregator. Petar et al.~\cite{petar2018graph} proposed the graph attention network (GAT) that learns to assign different aggregating weights to different nodes by leveraging the self-attention mechanism. Inspired by the above progress, we propose to learn primitive representations by global feature aggregations and use GCNs to transform primitive representations into visual text representations.

\section{Methodology}
\label{method}

In this section, we first describe methods for learning primitive representations and visual text representations, and then provide detailed structures of PREN and PREN2D.

\subsection{Primitive representation learning}

We propose learning primitive representations by using global feature aggregations over the coordinate space. In this way, primitive representations contain global information of the input image that is beneficial for the subsequent recognition process. Let $F\in \mathbb{R}^{d_0\times h\times w}$ be feature maps extracted by a CNN, where $h$, $w$, and $d_0$ are the height, width, and number of channels of $F$, respectively. The elements in feature maps are taken as the nodes of an undirected graph, i.e., we convert $F$ to a feature matrix $X\in \mathbb{R}^{m_0\times d_0}$, where $m_0=h\times w$. Let $n$ be the number of primitive representations to learn, the feature aggregation process can be formulated as

\begin{align}
    Z_i &= f^{(i)}(X),\ i=1,2,...,n \label{eq:ori1} \\ 
    \bm{p}_i &= \bm{a}_i\cdot Z_i,\ i=1,2,...,n \label{eq:ori2}
\end{align}

where $\bm{p}_i\in \mathbb{R}^{1\times d}$ is the $i$-th primitive representation. $f^{(i)}(\cdot)$ is the mapping function of a sub-network that transforms $X\in \mathbb{R}^{m_0\times d_0}$ into a hidden representation $Z_i\in \mathbb{R}^{m\times d}$. $\bm{a}_i\in \mathbb{R}^{1\times m}$ is the aggregating weights of the $i$-th primitive representation. The $n$ primitive representations are concatenated as $P=[\bm{p}_1; \bm{p}_2; ...; \bm{p}_n]\in \mathbb{R}^{n\times d}$.

We propose two kinds of aggregation methods with different aggregating weights $\bm{a}_i\ (i=1,2,...,n)$, i.e., a pooling aggregator and a weighted aggregator.

\subsubsection{Pooling aggregator}

As shown in Fig.~\ref{fig2} (a), a global average pooling layer is used for feature aggregation, which is equivalent to setting $a_{ij}=\frac{1}{m},\ \forall j=1,2,...,m$ in Equ.~(\ref{eq:ori2}). The global average pooling has been proven effective for learning global information~\cite{lin2014network,hu2018squeeze}. In this way, the aggregating weights are shared by all samples to exploit intrinsic structural information from various scene text instances.

The function $f^{(i)}(\cdot)$ in Equ.~(\ref{eq:ori1}) is implemented as two convolultions that conduct on the original feature maps $F$. Each convolution has kernel size = 3 and stride = 2. The calculation of primitive representations can be formulated as

\begin{align}
    \bm{p}_i &= Pool(conv_2^{(i)}(\phi(conv_1^{(i)}(F)))) \label{eqpl}
\end{align}

where $\phi(\cdot)$ denotes an activation function. Different from the pooling aggregator used in GraphSAGE~\cite{hamilton2017inductive}, we use additional convolutional layers before the pooling layer to better learn spatial information of scene text images.

\subsubsection{Weighted aggregator}

Due to the diversity of text instances in natural scene images, it is also important to learn sample-specific information. Therefore, we propose learning aggregating weights from input features dynamically.

\begin{figure*}[ht]
\centering
\includegraphics[width=.95\textwidth]{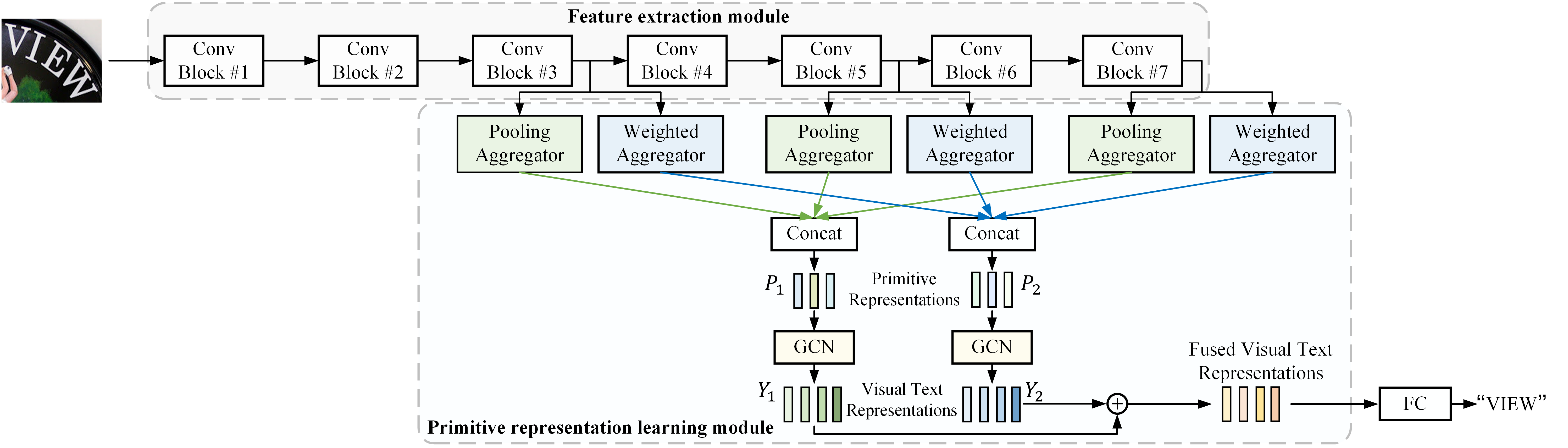}
\caption{The system framework of PREN that consists of a feature extraction module and a primitive representation learning module. Both pooling aggregators and weighted aggregators learn primitive representations $P_1$ and $P_2$ from feature maps. Visual text representations $Y_1$ and $Y_2$ are obtained from primitive representations $P_1$ and $P_2$ by two GCNs, and are summed into fused visual text representations for parallel decoding.}
\label{fig3}
\end{figure*}

As shown in Fig.~\ref{fig2} (b), a hidden representation $Z\in \mathbb{R}^{d\times h\times w}$ is obtained by a $3\times 3$ convolutional layer $conv_3(\cdot)$. Another $3\times 3$ convolutional layer $conv_4(\cdot)$ followed by a sigmoid activation function is used to convert input feature maps $F$ to $n$ heatmaps $H\in \mathbb{R}^{n\times h\times w}$. Aggregating weights $\bm{a}_i$ can be obtained by flattening the $i$-th heatmap $H_i$. Primitive representations can be calculated by a scale-dot product and summation operation, i.e., we have

\begin{align}
    Z &= \phi(conv_3(F)) \\
    H &= \sigma(conv_4(F)) \\
    \bm{p}_i &= \sum_{x=1}^h \sum_{y=1}^w H_{i,x,y}Z_{x,y} \label{eq:first}
\end{align}

\subsection{Visual text representation generation}

Since primitive representations contain global information of the input image, textual information can be extracted from primitive representations. We propose to generate visual text representations through a linear combination of primitive representations followed by a fully-connected layer, which can be formulated as

\begin{align}
    Y &= \phi(BPW) \label{eq:vtrg}
\end{align}

where $P\in \mathbb{R}^{n\times d}$ denotes primitive representations. $B\in \mathbb{R}^{L\times n}$ is a coefficient matrix of the linear combination, and $L$ is a maximum decoding length. $W\in \mathbb{R}^{d\times d}$ is a weight matrix, and $\phi(\cdot)$ is an activation function.

Equ.~(\ref{eq:vtrg}) can be implemented by using a GCN~\cite{kipf2016semi}, where the coefficient matrix $B$ plays a similar role to an adjacency matrix. Since there is no explicit graph topology for primitive representations, $B$ is randomly initialized and learned in the training stage.

Each visual text representation $\bm{y}_i\ (i=1,2,...,L)$ is used to represent a character to be recognized. For text string shorter than $L$, the excess part of $Y$ corresponds to padding symbols.

\subsection{Primitive representation learning network}
\label{sect:pren}

\subsubsection{Overview of PREN}

As shown in Fig.~\ref{fig3}, PREN consists of a feature extraction module and a primitive representation learning module. Three pooling aggregators and three weighted aggregators are used to learn primitive representations from multiscale feature maps. Let $P_1$ and $P_2$ denote primitive representations learned by pooling aggregators and weighted aggregators, respectively. Visual text representations $Y_1$ and $Y_2$ are obtained by two GCNs and are summed into fused visual text representations $Y$.
A fully-connected layer is used to convert $Y$ into logits for parallel decoding.

\subsubsection{Feature extraction module}

We use EfficientNet-B3~\cite{tan2019efficientnet} as the feature extraction module, which consists of seven mobile inverted bottlenecks (MBConv blocks)~\cite{sharir2017expressive,tan2019mnasnet}, as marked by ``Conv Block \#1'' to ``Conv Block \#7'' in Fig.~\ref{fig3}.

We denote the feature maps output by the $i$-th convolutional block by $F_i$. To take advantage of multiscale features, feature maps $F_3$, $F_5$, and $F_7$, which are $1/8$, $1/16$, and $1/32$ the input image scale, are used as inputs for the primitive representation learning module.

\subsubsection{Primitive representation learning module}

For feature maps output by each selected convolutional block, both a pooling aggregator and a weighted aggregator are used to learn primitive representations. Let $d$ denote the number of channels of $F_7$ and $n$ be the number of primitive representations to learn. The output of each feature aggregator has the dimension of $\mathbb{R}^{n\times \frac{d}{3}}$. As shown in Fig.~\ref{fig3}, the outputs of the three pooling aggregators are concatenated as $P_1\in \mathbb{R}^{n\times d}$, and the outputs of the three weighted aggregators are concatenated as $P_2\in \mathbb{R}^{n\times d}$.

Two GCNs are used to generate visual text representations $Y_1$ and $Y_2$ from primitive representations $P_1$ and $P_2$ respectively. $Y_1$ and $Y_2$ are summed into fused visual text representations $Y$. The probability of each character is computed from $Y$ through a fully-connected layer followed by softmax. Therefore, the decoding process of PREN is fully parallel and efficient.


    
    

\subsection{Incorporating the 2D attention mechanism}

The visual text representations output by PREN are also flexible to integrate into attention-based encoder-decoder models to alleviate the misalignment problem~\cite{cheng2017focusing,wang2019decoupled}. For attention-based methods, the alignment between texts and feature maps relies on previous decoded results. Since visual text representations are purely learned from visual features, they can provide global visual information that helps learn stable and accurate alignments.

\begin{figure}[ht]
\centering
\includegraphics[width=.5\textwidth]{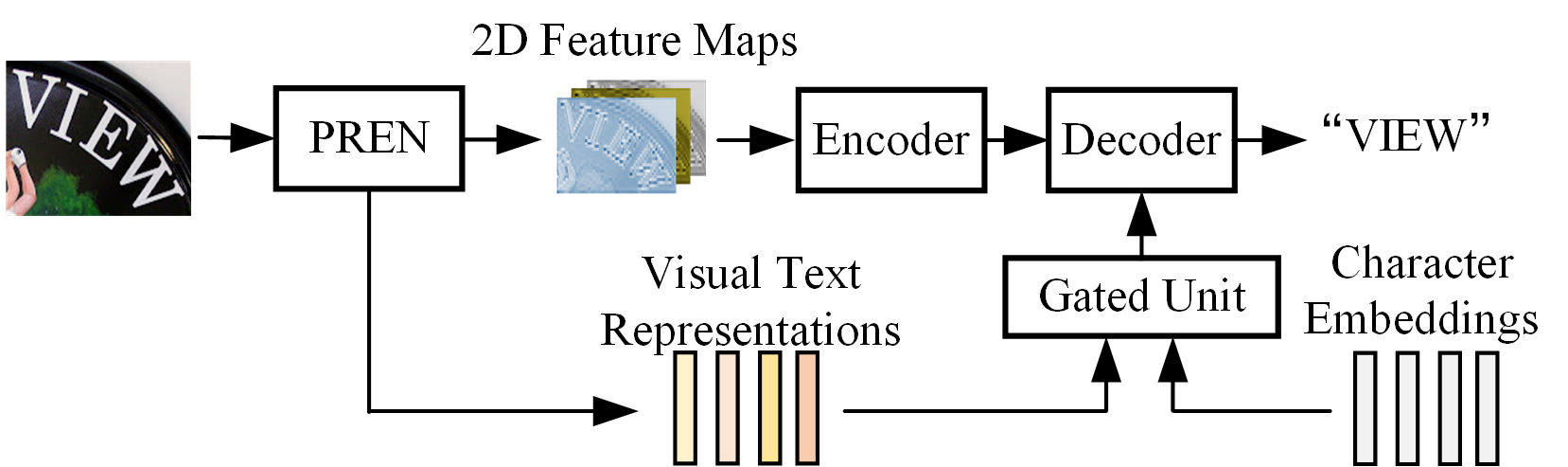}
\caption{Illustration of PREN2D. At each decoding step $t$, the $t$-th character embedding is combined with the $t$-th visual text representation by a gated unit.}
\label{fig35}
\end{figure}

Based on the above analysis, we construct PREN2D by combining PREN and a baseline model with the 2D attention mechanism. As shown in Fig.~\ref{fig35}, the feature extraction module is shared by both PREN and the encoder-decoder module based on a modified Transformer model~\cite{vaswani2017attention}. Visual text representations output by PREN are used to augment character embeddings of ground truth texts in the training stage or previous decoded texts in the inference stage, which can provide global guidance in the encoder-decoder attention calculation in the modified Transformer model.

For the feature extraction module, the outputs of the final convolutional block $F_7$ are upsampled and added to $F_5$, and the results are unsampled again and added to $F_3$. The obtained 2D feature maps $F\in \mathbb{R}^{d\times h\times w}$ have the same resolution as $F_3$ and the same number of channels as $F_7$.

In the original Transformer model, the encoder and decoder have $N=6$ identical blocks. In our model, the encoder and decoder are simplified to have $N=2$ identical blocks. For the encoder, we propose a modified self-attention mechanism that can be formulated as

\begin{align}
    q_i &= f(\mathcal{N}(f_i))\cdot W_Q \label{eqy1} \\
    k_j &= g(\mathcal{N}(f_j))\cdot W_K \label{eqy2} \\
    \alpha_{ij} &= softmax(\frac{1}{\sqrt{d}}q_ik_j^T) \\
    v_i &= \sum_{j=1}^m \alpha_{ij}x_jW_V
\end{align}

where $f_i\in \mathbb{R}^{1\times d}\ (i=1,2,...,m)$ is the $i$-th element in feature maps $F$, and $m=h\times w$. $W_Q, W_K, W_V\in \mathbb{R}^{d\times d}$ are three learnable matrices with respect to queries, keys and values. $\mathcal{N}(f_i)$ denotes spatially adjacent elements of $i$. $f(\mathcal{N}(f_i))$ and $g(\mathcal{N}(f_j))$ are implemented as two $3\times 3$ convolutional layers. In this way, the encoder can better model local spatial relationships during the computation of the attention weight $\alpha_{ij}$.

A Transformer decoder~\cite{vaswani2017attention} is used for text transcription. We use a gated unit to combine visual text representations and character embeddings. Let $Y$ and $E$ denote visual text representations and character embeddings, respectively. $V$ and $O$ are encoder outputs and decoder outputs, respectively. Formally, the calculation process of the decoder is

\begin{align}
    &z = \sigma([Y,E]\cdot W_z) \\
    &E'= z\odot Y + (1-z)\odot E \\
    &O = f_{dec}(E', V)
\end{align}

where $[\cdot]$ refers to concatenation, $W_z$ is a learnable weight matrix, $\odot$ denotes element-wise product, and $f_{dec}(\cdot)$ is the mapping function of the decoder.

\subsection{Training and inference}

Both PREN and PREN2D can be trained end-to-end with cross-entropy between the prediction and ground truth. The ground truth is generated by adding an ending symbol \textlangle eos\textrangle\ after the last character and expanded to a maximum decoding length with padding symbols \textlangle pad\textrangle. Let $l$ denote the length of the original text, and the loss is calculated according to

\begin{align}
    \mathcal{L}=-\sum_{t=1}^{l+1}logp(g_t|I)
\end{align}

where $I$ refers to the input image, $g_t\ (t=1,2,...,l)$ is the $t$-th character, and $g_{l+1}$ is the ending symbol \textlangle eos\textrangle. Padding symbols \textlangle pad\textrangle\ are ignored during the loss computation.

In the inference stage, PREN predicts the whole text at one step, while PREN2D recognizes characters recursively. The presence of the first ending symbol \textlangle eos\textrangle\ in the decoding results indicates the end of decoding.

\section{Experiments}
\label{sect:exp}

We conduct both English and Chinese scene text recognition experiments. For English scene text recognition, we compare our method with previous state-of-the-art methods and conduct a series of ablation studies to explore the effect of each part of our models. For Chinese scene text recognition, we evaluate the performance of our method on a multi-oriented text recognition task.

\begin{table*}[t]
\centering
\small
\caption{Word recognition accuracy (\%) across methods and datasets. MJ, ST, Char, and Add denote MJSynth~\cite{jaderberg2014synthetic}, SynthText~\cite{gupta2016synthetic}, character bounding boxes, and additional training data, respectively. The method with the * symbol had its results reported in Baek et al.~\cite{baek2019wrong}, where a reimplemented model is trained on MJ+ST. The best results of models trained on MJ+ST are marked in \textbf{bold}.}
\begin{tabular}{l l| c c c c| c c c}
\toprule
\multirow{2}{*}{Model} & \multirow{2}{*}{Training data} & \multicolumn{4}{|c|}{Regular Test Datasets} & \multicolumn{3}{c}{Irregular Test Datasets} \\
 &  & IIIT5k & SVT & IC03 & IC13 & IC15 & SVTP & CUTE \\
\midrule
Mask TextSpotter (Liao et al.)~\cite{liao2019mask} & MJ+ST+Char & 95.3 & 91.8 & 95.0 & 95.3 & 78.2 & 83.6 & 88.5 \\
SAR (Li et al.)~\cite{li2019show} & MJ+ST+Add & 95.0 & 91.2 & - & 94.0 & 78.8 & 86.4 & 89.6 \\
SCATTER (Litman et al.)~\cite{litman2020scatter} & MJ+ST+Add & 93.7 & 92.7 & 96.3 & 93.9 & 82.2 & 86.9 & 87.5 \\
\midrule
\midrule
CRNN (Shi et al.)~\cite{shi2016end,baek2019wrong}* & MJ+ST & 82.9 & 81.6 & 93.1 & 89.2 & 64.2 & 70.0 & 65.5  \\ 
AON (Cheng et al.)~\cite{cheng2018aon} & MJ+ST & 87.0 & 82.8 & 91.5 & - & 68.2 & 73.0 & 76.8 \\ 
DAN (Wang et al.)~\cite{wang2019decoupled} & MJ+ST & 94.3 & 89.2 & 95.0 & 93.9 & 74.5 & 80.0 & 84.4 \\
ASTER (Shi et al.)~\cite{shi2018aster} & MJ+ST & 93.4 & 89.5 & 94.5 & 91.8 &  76.1 & 78.5 & 79.5 \\
SE-ASTER (Qiao et al.)~\cite{qiao2020seed} & MJ+ST & 93.8 & 89.6 & - & 92.8 &  80.0 & 81.4 & 83.6 \\
AutoSTR (Zhang et al.)~\cite{zhang2020autostr} & MJ+ST & 94.7 & 90.9 & 93.3 & 94.2 & 81.8 & 81.7 & - \\
RobustScanner (Yue et al.)~\cite{yue2020robustscanner} & MJ+ST & 95.3 & 88.1 & - & 94.8 & 77.1 & 79.5 & 90.3 \\
SRN (Yu et al.)~\cite{yu2020towards} & MJ+ST & 94.8 & 91.5 & - & 95.5 & 82.7 & 85.1 & 87.8 \\ \midrule
CNN-LSTM-CTC & MJ+ST & 92.0 & 89.8 & 93.1 & 93.9 & 76.7 & 80.6 & 80.9 \\
PREN & MJ+ST & 92.1 & 92.0 & 94.9 & 94.7 & 79.2 & 83.9 & 81.3 \\
Baseline2D & MJ+ST & 95.4 & 93.4 & 95.4 & 95.9 & 81.9 & 86.0 & 89.9 \\
PREN2D & MJ+ST & \textbf{95.6} & \textbf{94.0} & \textbf{95.8} & \textbf{96.4} & \textbf{83.0} & \textbf{87.6} & \textbf{91.7} \\
\bottomrule
\end{tabular}
\label{tab6}
\end{table*}

\subsection{English scene text recognition}

\subsubsection{Experimental setup}
For English scene text recognition, our models are trained on two commonly used public synthetic scene text datasets, i.e., MJSynth (MJ)~\cite{jaderberg2014synthetic} and SynthText (ST)~\cite{gupta2016synthetic}. The model performance is tested on seven public real scene text datasets: IIIT5k-Words (IIIT5k)~\cite{mishra2012scene}, Street View Text (SVT)~\cite{wang2011end}, ICDAR 2003 (IC03)~\cite{lucas2005icdar}, ICDAR 2013 (IC13)~\cite{karatzas2013icdar}, ICDAR 2015 (IC15)~\cite{karatzas2015icdar}, SVT-Perspective (SVTP)~\cite{quy2013recognizing}, and CUTE80 (CUTE)~\cite{risnumawan2014robust}. There are various divisions for test sets of IC13 and IC15. We follow the protocol of Yu et al.~\cite{yu2020towards} where the IC13 test set consists of 857 images and the IC15 test set contains 1811 images. 

For ablation studies, all models are trained for three epochs. The learning rate of the first two epochs is set to 0.5 and decreased to 0.1 at the third epoch. When compared with other state-of-the-art methods, we continue to train the models for another five epochs. The learning rate is initialized to 0.1 and decreased to 0.01 and 0.001 at the third epoch and the fifth epoch, respectively. The training batch size is set to 128, and ADADELTA~\cite{zeiler2012adadelta} is adopted as the optimizer. Input images are normalized into $64\times 256$ pixels. The alphabet includes all case-insensitive alphanumerics. The number of primitive representations is 5. The maximum decoding length is set to 25 since the lengths of most common English words are less than 25. Word accuracy is used as the performance evaluation index.

\subsubsection{Comparison with state-of-the-art methods}
\label{sect:main}

The comparison of our models with previous state-of-the-art methods is shown in Table~\ref{tab6}. To better observe the performance gain of primitive representation learning, we also train a CTC-based model (CNN-LSTM-CTC) by replacing the CNN in the CRNN~\cite{shi2016end} with an EfficientNet-B3~\cite{tan2019efficientnet}, and train a baseline model with 2D attention mechanism (Baseline2D). Baseline2D has the same feature extraction module, encoder-decoder module, and training configurations as used in PREN2D. 

PREN achieves better recognition accuracy on all test sets than CNN-LSTM-CTC. By exploiting visual text representations, PREN2D outperforms Baseline2D on all test sets. In particular, accuracy gains of 1.1\%, 1.6\%, and 1.8\% are obtained on irregular text datasets IC15, SVTP, and CUTE, respectively. PREN2D also achieves higher accuracy than previous state-of-the-art models that are trained on the MJSynth~\cite{jaderberg2014synthetic} and SynthText~\cite{gupta2016synthetic} datasets. The recognition performance on both regular and irregular scene text image datasets shows the effectiveness of our method.

\subsubsection{Comparison of computation cost}
\label{sect:speed}

\begin{table}[ht]
\centering
\footnotesize
\caption{Comparison of the recognition speeds of various models. DL. Framework refers to deep-learning framework.}
\label{tab5}
\begin{tabular}{lccc}
\toprule
Model        & DL. Framework & NVIDIA GPU & Time   \\ \midrule
CNN-LSTM-CTC & \multirow{4}{*}{PyTorch}  & \multirow{4}{*}{Tesla V100}   & 23.6ms \\
PREN         &  &  & 22.7ms \\
Baseline2D &  &  & 61.6ms \\
PREN2D       &  &  & 67.4ms \\ \bottomrule
\end{tabular}
\end{table}

Table~\ref{tab5} shows the average recognition speeds of various models on a single image. 
PREN has a slightly higher recognition speed than CNN-LSTM-CTC. Compared with Baseline2D, the extra time consumption of PREN2D is only 5.8ms on average.




\subsubsection{Comparison of feature aggregation methods}

\begin{table}[ht]
\centering
\small
\setlength\tabcolsep{5pt}
\caption{Word accuracy (\%) of PRENs with various feature aggregation methods.}
\begin{tabular}{lccccc}
\toprule
Aggregator  & IIIT5k & IC03 & IC13 & SVTP & CUTE \\ \midrule
Pooling       & 91.1   & 93.5 & \textbf{94.7} & 79.2 & 77.4 \\
Weighted       & 90.0   & 92.0 & 93.2 & 79.5 & 77.4 \\
Pooling + Weighted & \textbf{91.8}   & \textbf{93.9} & \textbf{94.7} & \textbf{81.7} & \textbf{81.3} \\
\bottomrule
\end{tabular}
\label{tab2}
\end{table}

In this experiment, we study the effect of various feature aggregation methods. We compare PRENs with only pooling aggregators or only weighted aggregators, as well as with both pooling aggregators and weighted aggregators.

Table~\ref{tab2} lists the comparison results. Two phenomena can be observed. (1) The model with weighted aggregators has lower recognition accuracy than the model with pooling aggregators on regular text datasets (IIIT5k, IC03, and IC13), but achieves equal or higher recognition accuracy on irregular text datasets (SVTP and CUTE). (2) Combining the two aggregation methods can significantly improve recognition performance, especially on irregular text datasets.

\subsubsection{Comparison of various numbers of primitive representations}

\begin{figure}[htb]
\begin{center}
    \includegraphics[width=.95\linewidth]{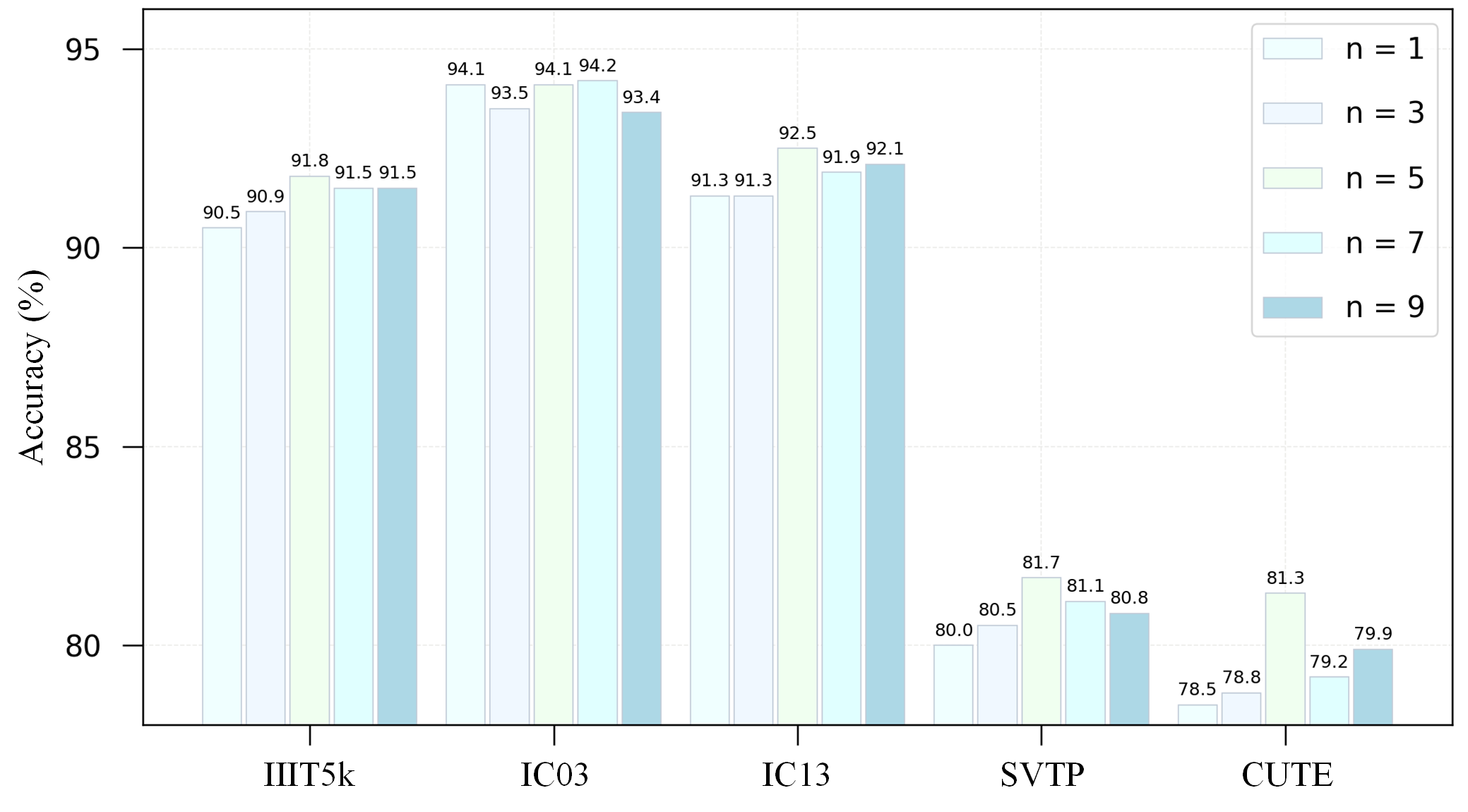}
\end{center}
   \caption{Comparison of PRENs with various numbers of primitive representations $n$.}
\label{fig:bar}
\end{figure}


Fig.~\ref{fig:bar} shows the comparison results of PRENs with various numbers of primitive representations. Too few or too many primitive representations will cause performance degradation. Learning five primitive representations achieves the best recognition performance on the IIIT5k, IC13, SVTP, and CUTE test sets.

\subsection{Chinese scene text recognition}

We further conduct a Chinese scene text recognition experiment. There are thousands of commonly used Chinese characters, and multi-oriented texts are common in Chinese scene images. Therefore, Chinese scene text recognition is a challenging task that can evaluate the robustness of scene text recognizers.

\subsubsection{Experimental setup}


For Chinese scene text recognition, our models are first trained on a self-built synthetic dataset, and then fine-tuned and tested on real samples. For the synthetic dataset, 1 million images are synthesized by following Gupta et al.~\cite{gupta2016synthetic} with the corpus collected from THUOCL~\cite{han2016thuocl}. The real samples with multiple orientations including horizontal, vertical, and skewed texts are selected from the RCTW~\cite{shi2017icdar2017} dataset. The training set for fine-tuning consists of 6000 images, and the test set includes 1000 images.

Since CTC-based models encode input images into feature sequences, a fixed normalized height is required for all images~\cite{choi2018simultaneous}. As a result, for CNN-LSTM-CTC, vertical text images are rotated 90 degrees first, and all images are normalized to $64\times 256$ pixels. In contrast, PREN, Baseline2D, and PREN2D can handle input images with multiple orientations. We divide the whole samples into horizontal and vertical subsets according to aspect ratios of original images. Horizontal and vertical text images are normalized into $64\times 256$ pixels and $256\times 64$ pixels, respectively. In the training stage, data of each training iteration is randomly taken from the horizontal subset or the vertical subset.

All models are trained on synthetic samples for 6 epochs and fine-tuned on real samples for 20 epochs. The learning rate is initialized to 0.5 and decreased to 0.1 at the sixth epoch. The character set contains 5658 characters.

\subsubsection{Comparison of different models}

\begin{table}[ht]
\centering
\small
\caption{Word accuracy (\%) of different models for multi-oriented Chinese scene text recognition.}
\label{tab:chinese}
\begin{tabular}{lccc}
\toprule
Model        & Horizontal & Vertical & Average \\ \midrule
CNN-LSTM-CTC & 53.4 & 64.8 & 59.1 \\
PREN         & 73.8 & 79.2 & 76.5 \\
Baseline2D   & 82.2 & 86.8 & 84.5 \\
PREN2D       & \textbf{82.6} & \textbf{87.4} & \textbf{85.0} \\ \bottomrule
\end{tabular}
\end{table}

Table~\ref{tab:chinese} shows the comparison results of different models. For CNN-LSTM-CTC, the rotation of vertical text images doubles the patterns to learn, while PREN can avoid this problem and achieve significantly higher accuracy. Baseline2D and PREN2D have better performance. One possible reason is that Chinese texts contain a lot of similar characters, thus the implicit language model learned by the encoder-decoder architecture is important for accurate recognition. Compared with Baseline2D, PREN2D achieves higher recognition accuracy on both horizontal and vertical test sets, which demonstrates the effectiveness of the proposed primitive representation learning method.

\subsection{Visualization and analysis}

\textbf{Qualitative comparison of different models.} Table~\ref{tab:sample} shows the qualitative comparison of different models. Irregular text images are challenging for CNN-LSTM-CTC. For PREN, errors are mainly caused by similar characters such as ``O'' and ``D''. Baseline2D suffers from the misalignment problem, e.g., the last character is repeatedly recognized twice for the third sample. PREN2D shows better robustness than the other three models.

\begin{table}[htb]
\centering
\caption{Qualitative comparison of different models. For Chinese texts, characters in Unicode form are also listed. Wrongly recognized characters are marked in red.}
\includegraphics[width=.95\linewidth]{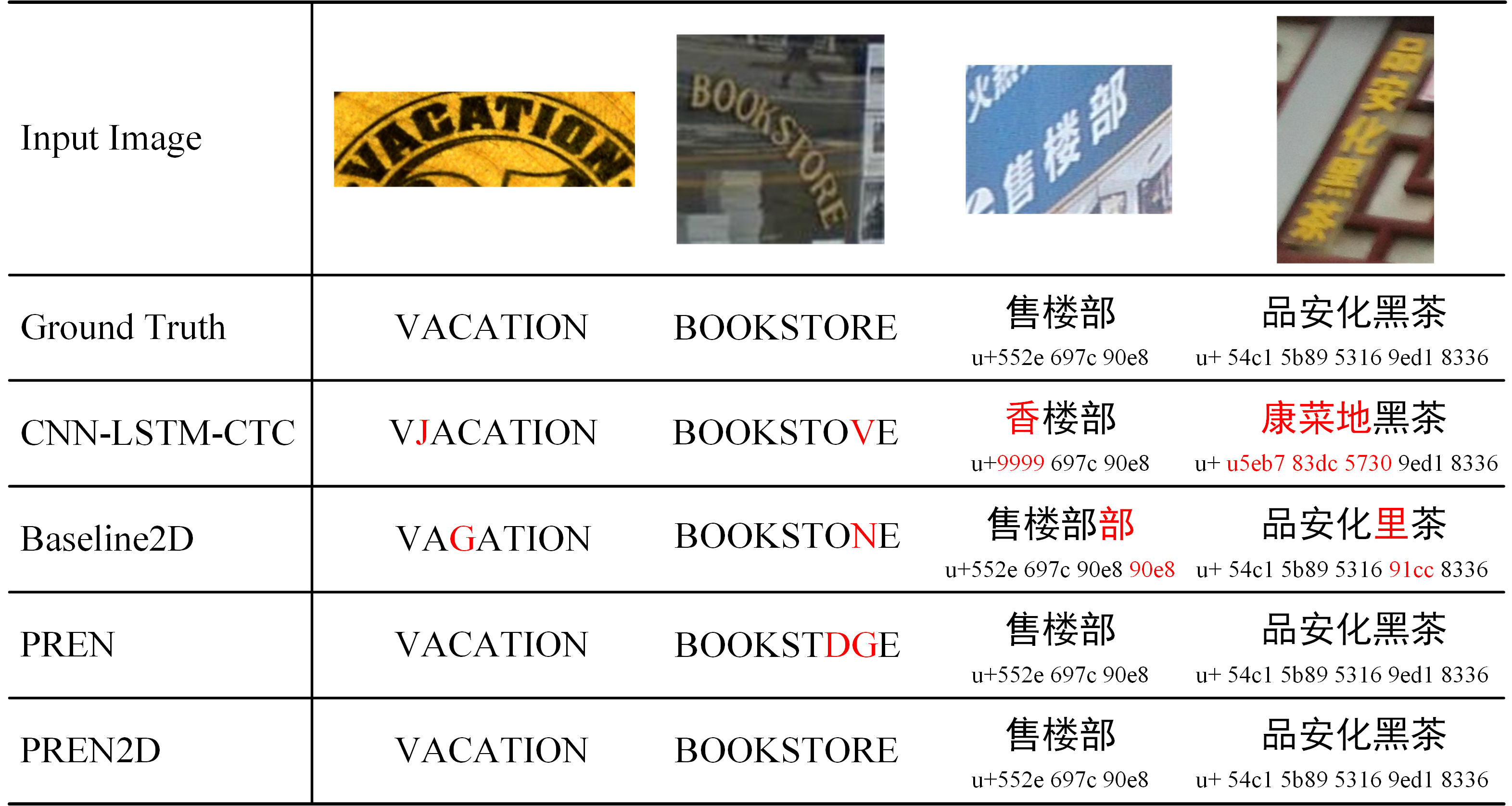}
\label{tab:sample}
\end{table}

\begin{figure}[htb]
\begin{center}
    \includegraphics[width=.95\linewidth]{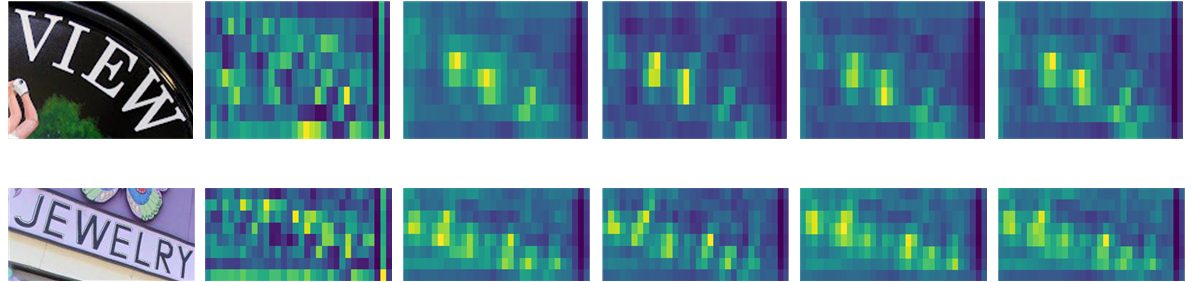}
\end{center}
   \caption{Visualization of heatmaps generated by the weighted aggregator that learns $n=5$ primitive representations.}
\label{fig5}
\end{figure}

\textbf{Visualization of different aggregators.} Fig.~\ref{fig5} shows the heatmaps generated by the weighted aggregator. The first heatmap has larger values in character boundary areas, while the other heatmaps focus on character center areas.

\begin{figure}[htb]
\begin{center}
    \includegraphics[width=.95\linewidth]{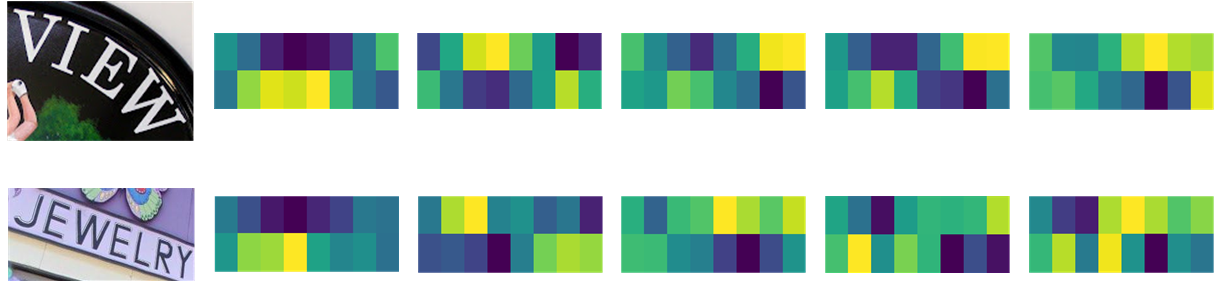}
\end{center}
  \caption{Feature maps before global average pooling of the pooling aggregator that learns $n=5$ primitive representations. Values are averaged in the channel dimension for visualization.}
\label{fig4}
\end{figure}

For the pooling aggregator, the feature maps before pooling (i.e., after $conv_2(\cdot)$ and before $Pool(\cdot)$ in Eq.~(\ref{eqpl})) show the contribution of each part of feature maps to primitive representations. As shown in Fig.~\ref{fig4}, for various input images, feature maps corresponding to the same primitive representation are similar, e.g., the first feature map generally has larger responses on the bottom parts and the second feature map focuses on the upper-left and lower-right parts.

The visualizations in Fig.~\ref{fig5} and Fig.~\ref{fig4} indicate that the pooling aggregator can learn common structural information from various text instances, and the weighted aggregator has a better ability to distinguish foreground and background areas.

\begin{figure}[htb]
\begin{center}
    \includegraphics[width=.95\linewidth]{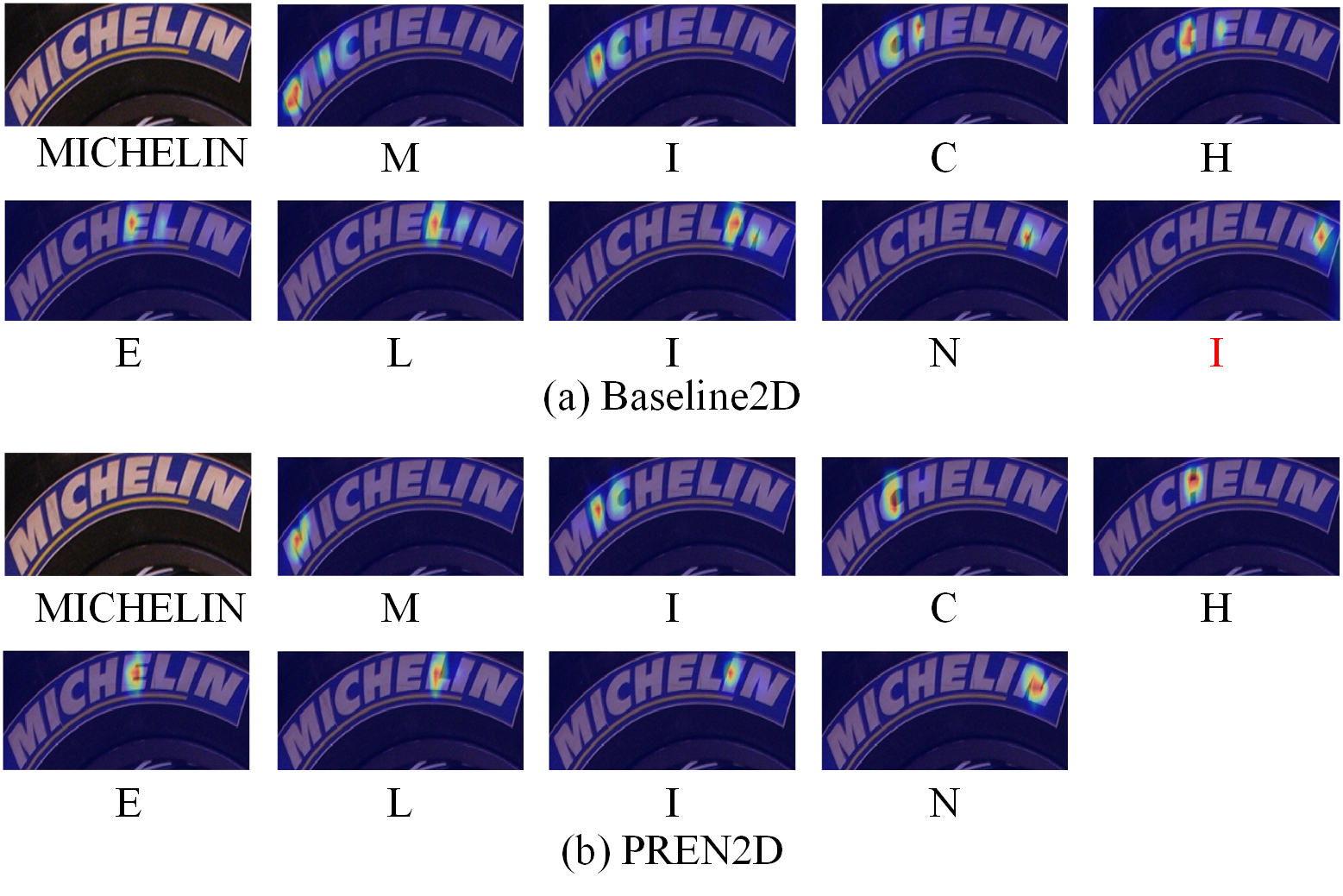}
\end{center}
   \caption{An example of attention scores generated by (a) the baseline model and (b) PREN2D. Texts under the input images are the ground truth, and the characters under attention maps are recognized results. The baseline model incorrectly recognizes ``MICHELIN'' as ``MICHELINI'', while PREN2D outputs correct results.}
\label{fig6}
\end{figure}

\textbf{Visualization of PREN2D.} Fig.~\ref{fig6} visualizes the attention scores generated by Baseline2D and PREN2D for an input image. By utilizing visual text representations, PREN2D can generate more accurate attention areas and alleviate incorrect alignments. For example, Baseline2D wrongly aligns the right part of ``N'' in the image with the character ``I''. In contrast, the attention map of PREN2D covers the central region of ``N'', in which the alignment is correct.

\section{Conclusion}

In this paper, we propose a primitive representation learning method for scene text recognition. Visual text representations generated from primitive representations can be either directly used for parallel decoding, or further integrated into a 2D-attention-based encoder-decoder framework to improve recognition performance. In future work, we will investigate more possible ways of learning primitive representations.

{\small
\bibliographystyle{ieee_fullname}
\bibliography{egbib}
}

\end{document}